# On-line Prediction with Kernels and the Complexity Approximation Principle

**Alex Gammerman**
Department of Computer Science,
Royal Holloway, University of London
Egham, Surrey, TW20 0EX, England
alex@cs.rhul.ac.uk

**Yuri Kalnishkan***
yura@cs.rhul.ac.uk

**Vladimir Vovk**
vovk@cs.rhul.ac.uk

## Abstract

The paper describes an application of Aggregating Algorithm to the problem of regression. It generalizes earlier results concerned with plain linear regression to kernel techniques and presents an on-line algorithm which performs nearly as well as any oblivious kernel predictor. The paper contains the derivation of an estimate on the performance of this algorithm. The estimate is then used to derive an application of the Complexity Approximation Principle to kernel methods.

## 1 INTRODUCTION

Papers [Vov90] and [VW98] introduce a technique called Aggregating Algorithm. This method applies to on-line learning settings. In the on-line model, a learner observes elements of a sequence one by one (some side information or *signals* may also be available) and, on each step, tries to predict the next yet unseen element. As the guesses differ from actual outcomes, the learner suffers loss, which accumulates over the iterations.

Suppose that there are other learners ('*experts*') performing along with ours and suppose that the predictions output by the experts are available to the learner before it produces its own. The problem of prediction with expert advice (see [CBFH+97, LW94]) is to predict *little* worse than any of the experts, i.e., to suffer the overall loss which is not much bigger than the loss of the best expert. Aggregating Algorithm is a method of merging experts' predictions in such a way as to provide a solution (which is optimal in some sense) for a wide range of problems of this type.

Aggregating Algorithm is not restricted to finite or even countable pools of experts and it can be applied to some class of prediction strategies, e.g., to all implementations of a given algorithm with different values of parameters. In [Vov98], Aggregating Algorithm is used to merge all linear functions which map the set of signals into the set of outcomes. By using the resulting strategy, which is called Aggregating Algorithm Regression or AAR, the learner performs nearly as well as it would do by sticking to any linear regression function. AAR is described in Subsect. 2.2.

However the applicability of the results from [Vov98] is limited since the paper considers only the simple linear regression. Linear regression is only able to determine the simplest dependencies in data. A more versatile tool is provided by kernel techniques. This paper generalizes the algorithm and the results from [Vov98] to classes of kernel strategies. It turns out to be possible to reduce AAR to a form that does not refer to the signal vectors except in scalar products; then the scalar products can be replaced by a kernel function. This is a widespread procedure sometimes referred to as the 'kernel trick' (see [SS02]). By applying this method, we derive the algorithm 'Kernel Aggregating Algorithm Regression' (KAAR) in Subsect. 3.1.

The main technical result of this paper is the bound on the performance of KAAR derived in Subsect. 3.2. The bound from [Vov98] is usually vacuous in the kernel setting since it includes a matrix of the same dimension as the space of signals ('feature space'). To obtain a result applicable in the kernel setting that bound is modified using a lemma from linear algebra.

Paper [Vov98] observes similarities between AAR and Ridge Regression; we discuss this matter in Subsect. 3.3.

In Sect. 4 we discuss how the bound obtained in this paper can be applied to implement the Complexity Approximation Principle introduced in [VG99]. This principle is an approach to model selection problem based on predictive complexity and generalizing the well-known Minimum Description Length principle.

---
* Corresponding author

## 2 PRELIMINARIES

### 2.1 Protocol

We consider the following on-line protocol. At each moment $t = 1, 2, \ldots$, the value of a *signal* $x_t \in X$ arrives. A prediction strategy $\mathfrak{A}$ observes $x_t$ and then outputs a prediction $\gamma_t \in \mathbb{R}$. Finally, the outcome $y_t \in [-Y, Y]$ arrives. This can be summarized in the following scheme:

**Protocol 1.**
```
FOR t = 1, 2, ...
    𝔄 reads x_t ∈ X;
    𝔄 outputs γ_t ∈ ℝ;
    𝔄 reads y_t ∈ [-Y, Y];
END FOR
```

The set $X$ is a *signal space*, which is assumed to be known to the strategy in advance. The bound $Y$ is a positive number and we do not make any assumptions on whether the strategy knows it. We will discuss this matter later.

The performance of $\mathfrak{A}$ is measured by the sum of squared discrepancies between the predictions and the outcomes. We say that on trial $t$ the strategy $\mathfrak{A}$ suffers loss $(y_t - \gamma_t)^2$. The losses incurred over several trials sum up to the overall loss. Thus after $T$ trials, the total loss of $\mathfrak{A}$ is

$$\text{Loss}_\mathfrak{A}\left((x_1, y_1), (x_2, y_2), \ldots, (x_T, y_T)\right)$$
$$= \sum_{t=1}^{T} (y_t - \gamma_t)^2 \ .$$

Informally, the goal of $\mathfrak{A}$ is to learn the dependency between $x$s and $y$s 'on the fly'. While studying, $\mathfrak{A}$ can make incorrect assumptions on the nature of the dependency and thus suffers loss.

### 2.2 Linear predictors

An important case of $X = \mathbb{R}^n$ is considered in [Vov01a]. That paper suggests the algorithm AAR ('Aggregating Algorithm Regression'). AAR receives a positive parameter $a$ and the dimension $n$ and operates as follows:

**Protocol 2 (AAR).**
```
A := aI;
b := 0;
FOR t = 1, 2, ...
    read x_t;
    A := A + x_t x_t';
    output γ_t := b' A^{-1} x_t;
    read y_t;
    b := b + y_t x_t;
END FOR
```

Here $A$ is an $n \times n$ real matrix and $I$ is the $n \times n$ unit matrix (note that in different contexts $I$ denotes matrices of different size; we do not include a reference to the dimension). When speaking about a vector $x \in \mathbb{R}^n$, we assume by default that it is a column vector; this applies to $x_t$ in the description of the protocol as well as to $b \in \mathbb{R}^n$. By $M'$ we denote the transpose of a matrix $M$.

Note that the inversion of $A$ can be performed for any positive $a$. Indeed, on trial $T$ we have $A = A_T = aI + \sum_{t=1}^{T} x_t x_t'$. The sum $\sum_{t=1}^{T} x_t x_t'$ is a symmetric and positive semidefinite matrix while the addition of $aI$ makes the sum $A_T$ positive definite.

The main property of AAR is that it performs little worse than any *oblivious linear predictor*. By the latter we mean a strategy that predicts $\theta' x_t$ on every trial $t$, where $\theta \in \mathbb{R}^n$ is some fixed vector. The set of all oblivious linear predictors may be identified with $\mathbb{R}^n$.

**Proposition 1 ([Vov01a]).** *For every positive integer $n$, every sequence*

$$S = ((x_1, y_1), (x_2, y_2), \ldots, (x_T, y_T))$$
$$\in (\mathbb{R}^n \times [-Y, Y])^* \ ,$$

*and every $a > 0$, the estimate*

$$\text{Loss}_{AAR}(S) \leq \inf_{\theta \in \mathbb{R}^n} \left(\text{Loss}_\theta(S) + a\|\theta\|^2\right)$$
$$+ Y^2 \ln \det \left(I + \frac{1}{a} \sum_{t=1}^{T} x_t x_t'\right) \quad (1)$$

*holds.*

By $\|\theta\|$ we denote the quadratic norm $\sqrt{\theta'\theta}$ of a vector $\theta$ and $P^*$ denotes the set of all finite strings of elements from a set $P$ (i.e., $P^* = \bigcup_{n=1}^{\infty} P^n$).

### 2.3 Kernel methods

A kernel on a set $X$ is a function $\mathcal{K} : X^2 \to \mathbb{R}$ which is symmetrical and positive semidefinite, i.e.,

(i) for every $x_1, x_2 \in X$, we have $\mathcal{K}(x_1, x_2) = \mathcal{K}(x_2, x_1)$ and

(ii) for every positive integer $t$ and every sequence $x_1, x_2, \ldots, x_t \in X$, the matrix $\mathbf{K} = (\mathcal{K}(x_i, x_j))_{i,j}$ $(i, j = 1, 2, \ldots, t)$ is positive semidefinite.

The kernels can be characterized in the following way (see [Aro50, Wah90]).

**Proposition 2.** *Suppose that a function $\mathcal{K}$ is defined on $X^2$. Then $\mathcal{K}$ is a kernel if and only if there exists a Hilbert space $H$ with a scalar product $\langle \cdot, \cdot \rangle$ and a mapping $F : X \to H$ such that for every $x_1, x_2 \in X$ the equality $\mathcal{K}(x_1, x_2) = \langle F(x_1), F(x_2) \rangle$ holds.*

In this paper we need a slightly elaborated version of this. It is essential for us to obtain a separable Hilbert space, i.e., a Hilbert space isomorphic to $l_2$. Let us give the following definition.

**Definition 1.** A kernel $\mathcal{K}$ on a space $X$ is *separably implementable* if there is a separable Hilbert space $H$ with a scalar product $\langle \cdot, \cdot \rangle$ and a mapping $F: X \to H$ such that for every $x_1, x_2 \in X$ the equality $\mathcal{K}(x_1, x_2) = \langle F(x_1), F(x_2) \rangle$ holds.

By slightly modifying the proof from [Wah90], one may obtain the following sufficient condition.

**Proposition 3 ([Kre63]).** *If a kernel $\mathcal{K}: X \times X \to \mathbb{R}$ is defined over a separable topological space $X$, continuous in each of its variables, and $\mathcal{K}(x, x)$ is continuous as a function of one variable, then the kernel is separably implementable.*

## 3　KERNEL PREDICTORS

### 3.1　The kernel version of AAR

The protocol for AAR may be rewritten in a form that only includes $x_t$ in mutual scalar products:

**Lemma 1.** *On trial $t$, AAR makes a prediction $\gamma_t = \widetilde{\mathcal{Y}}'(aI + \widetilde{\mathbf{K}})^{-1}\widetilde{\mathbf{k}}(x)$, where $\widetilde{\mathcal{Y}} = (y_1, y_2, \ldots, y_{t-1}, 0)' \in \mathbb{R}^t$ is the vector of outcomes appended by zero, $\widetilde{\mathbf{k}}(x_t) = \left(x_1'x_t, x_2'x_t, \ldots, x_{t-1}'x_t, x_t'x_t\right)' \in \mathbb{R}^t$, and $\widetilde{\mathbf{K}}$ is the $t \times t$ matrix of scalar products $(x_i'x_j)_{i,j}$ $(i, j = 1, 2, \ldots, t)$.*

*Proof.* Let $M$ be the matrix consisting of column vectors $x_1, x_2, \ldots, x_t$, i.e., $M = (x_1, x_2, \ldots, x_t)$. On trial $t$, AAR outputs the prediction $\gamma_t = b'A^{-1}x_t$, where $b = b_t = \sum_{i=1}^{t-1} y_i x_i = M\widetilde{\mathcal{Y}}$ and $A = A_t = \sum_{i=1}^{t} x_i x_i' = MM'$. One can see that $\widetilde{\mathbf{K}} = M'M$ and $\widetilde{\mathbf{k}}(x_t) = Mx_t$ so it suffices to prove that

$$\widetilde{\mathcal{Y}}'M'(aI + MM')^{-1}x_t = \widetilde{\mathcal{Y}}'(aI + M'M)^{-1}M'x_t$$

or $M'(aI + MM')^{-1} = (aI + M'M)^{-1}M'$. Since both the matrixes $aI + MM'$ and $aI + M'M$ are positive definite and thus nonsingular, the last equality is equivalent to $(aI + M'M)M' = M'(aI + MM')$, which follows from the distributivity of matrix multiplication. □

This lemma motivates the extension of AAR to a wider class of signal spaces. Suppose we are given an arbitrary signal space $X$ with a kernel $\mathcal{K}$ on it. Let us define the algorithm KAAR ('Kernel Aggregating Algorithm Regression'). It accepts a positive real parameter $a$ and works according to the following protocol.

**Protocol 3 (KAAR).**
```
FOR t = 1, 2, ...
   read x_t;
```
$\quad\widetilde{\mathcal{Y}} := (y_1, y_2, \ldots, y_{t-1}, 0)';$
$\quad\widetilde{\mathbf{k}}(x_t) := \big(\mathcal{K}(x_1, x_t), \mathcal{K}(x_2, x_t), \ldots,$
$\qquad\qquad\qquad \mathcal{K}(x_{t-1}, x_t), \mathcal{K}(x_t, x_t)\big)';$
$\quad\widetilde{\mathbf{K}} := (\mathcal{K}(x_i, x_j))_{i,j}, \ i, j = 1, 2, \ldots, t;$
```
   output γ_t := Ỹ'(aI + K̃)^{-1}k̃(x);
   read y_t;
END FOR.
```

### 3.2　Upper bound

Like AAR, KAAR has an optimality property. KAAR performs little worse than any *oblivious kernel predictor*. An oblivious kernel predictor is a linear combination of functions of the form $\mathcal{K}(x, \cdot)$ or, formally, a strategy which is defined by a finite sequence of pairs

$$\mathfrak{S} = ((c_1, z_1), (c_2, z_2), \ldots, (c_n, z_n)) \in (\mathbb{R} \times X)^* \quad (2)$$

and predicts $c_1\mathcal{K}(z_1, x_t) + c_2\mathcal{K}(z_2, x_t) + \ldots + c_n\mathcal{K}(z_n, x_t)$ on every signal $x_t \in X$. Obviously,

$$\text{Loss}_\mathfrak{S}\left((x_1, y_1), (x_2, y_2), \ldots, (x_T, y_T)\right)$$
$$= \sum_{t=1}^{T} \left(\sum_{i=1}^{n} c_i \mathcal{K}(z_i, x_t) - y_t\right)^2 .$$

The following theorem generalizes Proposition 1. Note that the proof of Proposition 1 given in [Vov01a] cannot be repeated directly since it involved the evaluation of an integral over the signal space $X = \mathbb{R}^n$.

**Theorem 1.** *Let $\mathcal{K}$ be a separably implementable kernel on a space $X$. Then for every $a > 0$ and every sequence*

$$S = ((x_1, y_1), (x_2, y_2), \ldots, (x_T, y_T))$$
$$\in (X \times [-Y, Y])^* ,$$

*the estimate*

$$\text{Loss}_{KAAR}(S)$$
$$\leq \inf_{\mathfrak{S}=((c_1, z_1), \ldots, (c_n, z_n)) \in (\mathbb{R} \times X)^*} \Bigg( \text{Loss}_\mathfrak{S}(S)$$
$$+ a \sum_{i,j=1}^{n} c_i c_j \mathcal{K}(z_i, z_j) \Bigg)$$
$$+ Y^2 \ln \det \left(I + \frac{1}{a}\widetilde{\mathbf{K}}\right) \quad (3)$$

*holds, where $\widetilde{\mathbf{K}}$ is the $T \times T$ matrix $(\mathcal{K}(x_p, x_r))_{p,r}$, $p, r = 1, 2, \ldots, T$.*

*Proof.* We start with the special case $X = \mathbb{R}^m$ and $\mathcal{K}(z_1, z_2) = z_1'z_2$ for every $z_1, z_2 \in X$. In this case, (3) follows directly from (1). Indeed, a kernel predictor $\mathfrak{S} = ((c_1, z_1), (c_2, z_2), \ldots, (c_n, z_n))$ reduces to the linear predictor $\theta = \sum_{i=1}^{n} c_i z_i$ and the term $\sum_{i,j=1}^{n} c_i c_j \mathcal{K}(z_i, z_j)$ equals the squared quadratic norm of $\theta$. It remains to show that

$$\det\left(I + \frac{1}{a}\widetilde{\mathbf{K}}\right) = \det\left(I + \frac{1}{a}\sum_{t=1}^{T} x_t x_t'\right) \ .$$

This follows from the lemma below. This lemma follows from the formulas of Schur and sometimes is attributed to Sylvester (see the survey [HS81]). The lemma may also be derived using the LU-decomposition. A short self-contained derivation of the lemma is given in Appendix A.

**Lemma 2.** *For every matrix $M$ the equality*

$$\det(I + M'M) = \det(I + MM')$$

*holds.*

The general case can be obtained by using finite-dimensional approximations. Without loss of generality, we may assume that there is $F : X \to l_2$ such that for every $z_1, z_2 \in X$ we have $\mathcal{K}(z_1, z_2) = \langle F(z_1), F(z_2) \rangle$, where $\langle \alpha, \beta \rangle = \sum_{i=1}^{\infty} \alpha_i \beta_i$ is the scalar product of $\alpha = (\alpha_1, \alpha_2, \ldots), \beta = (\beta_1, \beta_2, \ldots) \in l_2 = \{\alpha = (\alpha_1, \alpha_2, \ldots) \mid \sum_{i=1}^{\infty} \alpha_i^2 \text{ converges }\}$.

Let us consider the sequence on subspaces $R_1 \subseteq R_2 \subseteq \ldots \subseteq l_2$, where $R_s = \{\alpha = (\alpha_1, \alpha_2, \ldots) \mid \forall v > s : \alpha_v = 0\} \subseteq l_2$. The set $R_s$ may be identified with $\mathbb{R}^s$. Let $p_s : l_2 \to R_s$ be the projection $p_s((\alpha_1, \alpha_2, \ldots)) = (\alpha_1, \alpha_2, \ldots, \alpha_s, 0, 0, \ldots)$, $F_s : X \to R_s$ be $F_s = p_s(F)$, and $\mathcal{K}_s$ be given by $\mathcal{K}_s(z_1, z_2) = \langle F_s(z_1), F_s(z_2) \rangle$, where $z_1, z_2 \in X$.

Inequality (3) holds for $\mathcal{K}_s$ since $R_s$ has a finite dimension. If (3) is violated, then its counterpart with some large $s$ is violated too and this observation completes the proof.

□

### 3.3 Comparison with Ridge Regression

The formulae from Protocol 3 are apparently reminiscent of Ridge Regression and indeed there are parallels.

Ridge Regression (RR) works as follows. Suppose we are given pairs $(x_1, y_2), (x_2, y_2), \ldots, (x_{t-1}, y_{t-1})$ and the goal is to predict the outcome for the signal $x_t$. If we are given a kernel $\mathcal{K}$ and a parameter (called ridge) $a > 0$, RR suggests predicting $r_t = \mathcal{Y}'(aI + \mathbf{K})^{-1}\mathbf{k}(x_t)$, where $\mathcal{Y} = (y_1, y_2, \ldots, y_{t-1})'$, $\mathbf{k}(x_t) = (\mathcal{K}(x_1, x_t), \mathcal{K}(x_2, x_t), \ldots, \mathcal{K}(x_{t-1}, x_t))'$, and $\mathbf{K} = (\mathcal{K}(x_i, x_j))_{i,j}$, $i, j = 1, 2, \ldots, (t-1)$. For a more detailed treatment of RR, refer to [CST00] or [SGV98].

It was observed in [Vov01a] that the prediction of AAR on a signal $x_t$ corresponds to the prediction of RR (with the scalar product as the kernel) which has received the same pairs of signals and outcomes and an additional pair $(x_t, 0)$. The same remains true for KAAR with an arbitrary kernel.

Paper [Vov01a] also mentions the following relation between AAR and Ridge Regression. Suppose that $r_T$ is the prediction output by Ridge Regression on the training set $(x_1, y_1), (x_2, y_2), \ldots, (x_{T-1}, y_{T-1})$ and input $x_T$, while $\gamma_T$ is the prediction output by KAAR on trial $T$ after getting the same sequence of signals and outcomes. Then $r_T$ and $\gamma_t$ satisfy the equality

$$\gamma_t = \frac{r_T}{1 + x_T' A_{T-1}^{-1} x_t} \ , \qquad (4)$$

where $A_{T-1} = aI + \sum_{t=1}^{T-1} x_t x_t'$.

In the kernel case the dependency between KAAR and RR is given by the following formulae. If $\gamma_t$ is as in Protocol 3, and $r_t$, $\mathbf{K}$, and $\mathbf{k}$ are as above, then

$$\gamma_t = \frac{a\mathcal{Y}'(\mathbf{K} + aI)^{-1}\mathbf{k}(x_t)}{\mathcal{K}(x_t, x_t) + a - (\mathbf{k}(x_t))'(\mathbf{K} + aI)^{-1}\mathbf{k}(x_t)} \qquad (5)$$

$$= \frac{ar_t}{\mathcal{K}(x_t, x_t) + a - (\mathbf{k}(x_t))'(\mathbf{K} + aI)^{-1}\mathbf{k}(x_t)} \ . \quad (6)$$

This representation follows from the formulae for inverting a partitioned matrix (see [PTVF94], Section 2.7) applied to $\widetilde{\mathbf{K}}$; more details may be found in Appendix B.

## 4 COMPLEXITY APPROXIMATION

Complexity Approximation Principle (CAP) was first introduced in [VG99]. CAP is an approach to the model selection problem. It deals with the following problem. Suppose that we are given a sequence $S = ((x_1, y_1), (x_2, y_2), \ldots, (x_T, y_T)) \in (X \times [-Y, Y])^*$ and a pool of prediction strategies (the strategies may be oblivious, i.e., just mappings $X \to [-Y, Y]$). We know how each of the strategies behaves on $S$. The task is to choose one strategy to use "in the future", i.e., a strategy that describes the data best.

We are interested in the squared discrepancy in this paper. One possible solution is to choose the least squares estimate, i.e., to minimize the value $\text{Loss}_{\mathfrak{A}}(S)$ over $\mathfrak{A}$. However this may lead to overfitting.

The situation is very similar to that with model selection in statistics. Least squares is the direct counterpart of maximum likelihood. One of the ways of combating overfitting problem in statistical modeling is to use the Minimum Description Length (MDL) principle, which suggests balancing "goodness-of-fit" of a hypothesis against its Kolmogorov complexity. CAP, suggested in [VG99], is a way of extending MDL to a wider range of situations including the case of squared discrepancy. CAP relies on the concept of predictive complexity.

### 4.1 Predictive Complexity

Consider a sequence of signals and outcomes $S = ((x_1, y_1), (x_2, y_2), \ldots, (x_T, y_T))$. One may want to define predictive complexity of a sequence as the loss of an optimal prediction strategy. Unfortunately, there is no such thing as a prediction strategy optimal in some natural sense; a simple diagonalization argument shows that every strategy is greatly outperformed by some other strategy on some sequence. However this obstacle can be avoided. Let us extend the class of prediction strategies to include certain quasi-strategies.

We say that $L : (X \times [-Y, Y])^* \to [0, \infty]$ is a superloss process if

- $L(\Lambda) = 0$, where $\Lambda$ is the empty string;

- for every sequence $((x_1, y_1), (x_2, y_2), \ldots, (x_T, y_T)) \in (X \times [-Y, Y])^*$ and every $x_{T+1} \in X$ there is $\gamma \in \mathbb{R}$ such that the inequality
$$L((x_1, y_1), (x_2, y_2), \ldots, (x_T, y_T), (x_{T+1}, y))$$
$$\geq L((x_1, y_1), (x_2, y_2), \ldots, (x_T, y_T)) + (y - \gamma)^2$$
holds for every $y \in [-Y, Y]$;

- $L$ is semicomputable from above, i.e., there is a computable sequence of computable functions $L_i(X \times [-Y, Y])^*$, $i = 1, 2, \ldots$, such that for every $S \in (X \times [-Y, Y])^*$ we have $L(S) = \inf_{i=1,2,\ldots} L_i(S)$.

Note that for every computable prediction strategy $\mathfrak{A}$ its loss $\text{Loss}_{\mathfrak{A}}$ is a superloss process.

A superloss process $L$ is called *universal* if it is the smallest up to an additive constant, i.e., if for every other superloss process $L'$ there is a constant $C > 0$ such that for every $S = ((x_1, y_1), (x_2, y_2), \ldots, (x_T, y_T)) \in (X \times [-Y, Y])^*$ the inequality $L(S) \leq L'(S) + C$ holds. We may pick a universal superloss process and call it predictive complexity $\mathcal{KP}^{X,Y}$. For every computable prediction strategy $\mathfrak{A}$ there is a constant $C > 0$ such that the inequality
$$\mathcal{KP}^{X,Y}(S) \leq \text{Loss}_{\mathfrak{A}}(S) + C$$
holds for all $S \in (X \times [-Y, Y])^*$.

It can be shown (see, e.g., [Vov01b]) that there exists (square loss) predictive complexity $\mathcal{KP}^{X,Y}$ (note that it depends on $X$ and $Y$). In fact, there is a general theory of relations between loss functions (we use the square loss $(y - \gamma)^2$) and predictive complexity; see [VW98] for more details. It can be shown that the concept predictive complexity is a generalization of Kolmogorov complexity; the square-loss complexity we are considering is an alternative to Kolmogorov complexity applicable to the case of square loss.

### 4.2 Complexity Approximation Principle

Let us fix a space of signals $X$ and a bound $Y > 0$ and let $\mathcal{KP} = \mathcal{KP}^{X,Y}$. It can be shown (see [VG99]) that there is a constant $C > 0$ such that for every prediction strategy $\mathfrak{A}$ and every sequence $S = ((x_1, y_1), (x_2, y_2), \ldots, (x_T, y_T)) \in (X \times [-Y, Y])^*$ we have

$$\mathcal{KP}(S) \leq \text{Loss}_{\mathfrak{A}}(S) + 2Y^2 \ln 2K(\mathfrak{A}) + C \ , \quad (7)$$

where K stands for prefix complexity (see [LV97] for a definition).

We can now formulate the Complexity Approximation Principle.

*CAP: Choose the prediction strategy that provides the best upper bound on the predictive complexity.*

Note that the right-hand side of (7) is only a special case of CAP. Any bound on predictive complexity can be used. Moreover, CAP can be applied to loss functions other than the square loss $(y - \gamma)^2$. The application of CAP to so called logarithmic loss gives the standard MDL, so CAP is really a generalization of MDL (see [VG99]).

### 4.3 Complexity Approximation for Kernel Methods

AAR is a computable strategy and therefore its loss provides an upper bound to $\mathcal{KP}$, namely, there is $C > 0$ such that for every $S = ((x_1, y_1), (x_2, y_2), \ldots, (x_T, y_T)) \in (X \times [-Y, Y])^*$ we have

$$\mathcal{KP}(S) \leq \text{Loss}_{AAR}(S) + C$$
$$\leq \inf_{\theta \in \mathbb{R}^n} \left( \text{Loss}_\theta(S) + a\|\theta\|^2 \right)$$
$$+ Y^2 \ln \det \left( I + \frac{1}{a} \sum_{t=1}^T x_t x_t' \right) + C \ . \quad (8)$$

Let us apply CAP to this estimate in order to choose $\theta$. CAP suggests choosing $\theta$ minimizing the sum $\text{Loss}_\theta(S) + a\|\theta\|^2$; we thus come to the conclusion that Ridge Regression is a CAP method. Note that here we speak about Ridge Regression as a batch method;

$\theta$ can only be obtained when all examples are available, i.e., in batch settings.

Theorem 1 generalizes this estimate and this conclusion to the case of an arbitrary (separably implementable and computable) kernel.

Suppose that we have a family of kernels $\mathcal{K}_m$ parameterized by $m = 1, 2, \ldots$, e.g., polynomial kernels $\mathcal{K}_m(x_1, x_1) = (1 + x_1' x_2)^m$, and we must choose $m$. Let $\mathfrak{A}_m$ be the prediction strategy implementing KAAR with the kernel $\mathcal{K}_m$. Applying (7) yields

$$\mathcal{KP}(S) \leq \min_{m=1,2,\ldots} \left( \text{Loss}_{\mathfrak{A}_m}(S) + \mathrm{K}(m) \right) + C$$

$$\leq \min_{m=1,2,\ldots} \left( \inf_{\mathfrak{S}=((c_1,z_1),\ldots,(c_n,z_n))\in(\mathbb{R}\times X)^*} \left( \text{Loss}_{\mathfrak{S}}(S) \right. \right.$$
$$\left. + a \sum_{i,j=1}^n c_i c_j \mathcal{K}_m(z_i, z_j) \right) + Y^2 \ln\det\left(I + \frac{1}{a}\widetilde{\mathbf{K}}_m\right)$$
$$\left. + (2Y^2 \ln 2)\mathrm{K}(m) \right) + C \ ,$$

where $\widetilde{\mathbf{K}}_m$ is the $T \times T$ matrix $(\mathcal{K}_m(x_p, x_r))_{p,r}$, $p, r = 1, 2, \ldots, T$. Let us minimize this expression over $\mathfrak{S}$ and over $m$ separately. Minimization over $\mathfrak{S}$ yields Ridge Regression; let $\text{Loss}_{RR,m}(S)$ be the loss of Ridge Regression with $\mathcal{K}_m$ on $S$ (recall the note above about batch settings). Let us use the bound $\mathrm{K}(m) \leq \log_2 m + 2 \log_2 \log_2 m + C$ on prefix complexity. Thus to obtain $m$ we should minimize the sum $\text{Loss}_{RR,m}(S) + Y^2 \ln\det\left(I + \frac{1}{a}\widetilde{\mathbf{K}}_m\right) + (2Y^2 \ln 2)(\log_2 m + 2 \log_2 \log_2 m)$.

The first term represents "goodness-of-fit" and it probably decreases as $m$ increases. The other terms balance this and we get a trade-off as a result.

**Acknowledgments**

The authors have been supported by the EPSRC through the grant GR/R46670 'Complexity Approximation Principle and Predictive Complexity: Analysis and Applications'. This research has also been partially supported by the NSF grant CCR-0325463 'New Directions in Predictive Learning: Rigorous Learning Machines', and the CLASS Center of Columbia University.

The authors would like to thank anonymous UAI reviewers for their in-depth comments on the paper.

# References

[Aro50] N. Aronszajn. Theory of reproducing kernels. *Transactions of the American Mathematical Society.*, 68:337–404, 1950.

[CBFH+97] N. Cesa-Bianchi, Y. Freund, D. Haussler, D. P. Helmbold, R. E. Schapire, and M. K. Warmuth. How to use expert advice. *Journal of the ACM*, 44(3):427–485, 1997.

[CST00] Nello Cristianini and John Shaw-Taylor. *An Introduction to Support Vector Machines.* Cambrige University Press, 2000.

[HS81] H. V. Henderson and S. R. Searle. On deriving the inverse of a sum of matrices. *SIAM Review*, 23(1), 1981.

[Kre63] M. Krein. Hermitian-positive kernels on homogeneous spaces. *American Mathematical Society Translations: Series 2*, 34:69–164, 1963.

[LV97] M. Li and P. Vitányi. *An Introduction to Kolmogorov Complexity and Its Applications.* Springer, New York, 1997.

[LW94] N. Littlestone and M. K. Warmuth. The weighted majority algorithm. *Information and Computation*, 108:212–261, 1994.

[PTVF94] William H. Press, Saul A. Teukolsky, William T. Vetterling, and Brian P. Flannery. *Numerical Recipies in C.* Cambrige University Press, 2nd edition, 1994.

[SGV98] C. Saunders, A. Gammerman, and V. Vovk. Ridge regression learning algorithm in dual variables. In *Proceedings of the 15th International Conference on Machine Learning*, pages 515–521, 1998.

[SS02] B. Schölkopf and A. J. Smola. *Learning with Kernels.* MIT Press, 2002.

[VG99] V. Vovk and A. Gammerman. Complexity approximation principle. *The Computer Journal*, 42(4):318–322, 1999.

[Vov90] V. Vovk. Aggregating strategies. In M. Fulk and J. Case, editors, *Proceedings of the 3rd Annual Workshop on Computational Learning Theory*, pages 371–383, San Mateo, CA, 1990. Morgan Kaufmann.

[Vov98] V. Vovk. Competitive on-line linear regression. In Michael I. Jordan, Michael J. Kearns, and Sara A. Solla, editors, *Advances in Neural Information Processing Systems*, volume 10. The MIT Press, 1998.

[Vov01a] V. Vovk. Competitive on-line statistics. *International Statistical Review*, 69(2):213–248, 2001.

[Vov01b] V. Vovk. Probability theory for the Brier game. *Theoretical Computer Science*, 261:57–79, 2001.

[VW98] V. Vovk and C. J. H. C. Watkins. Universal portfolio selection. In *Proceedings of the 11th Annual Conference on Computational Learning Theory*, pages 12–23, 1998.

[Wah90] G. Wahba. *Spline Models for Observational Data*. SIAM, Philadelphia, 1990.

## Appendix A: Proof of Lemma 2

Suppose than $M$ is an $n \times m$ matrix. Thus $I + MM'$ and $I + M'M$ are $n \times n$ and $m \times m$ matrixes, respectively. Without restricting the generality, we may assume that $n \geq m$ (otherwise we swap $M$ and $M'$). Let columns of $M$ be $m$ vectors $x_1, \ldots, x_m \in \mathbb{R}^n$.

We have $MM' = \sum_{i=1}^{n} x_i x_i'$. Let us see how the operator $MM'$ acts on a vector $x \in \mathbb{R}^n$. By associativity, $x_i x_i' x = (x_i' x) x_i$, where $x_i' x$ is a scalar. Therefore, if $U$ is the span of $x_1, x_2, \ldots, x_n$, then $MM'(\mathbb{R}^n) \subseteq U$ and $(I + MM')(\mathbb{R}^n) \subseteq U$ too. On the other hand, if $x$ is orthogonal to $x_i$, then $x_i x_i' x = (x_i' x) x_i = 0$. Hence $MM'(U^\perp) = 0$, where $U^\perp$ is the orthogonal complement to $U$ w.r.t. $\mathbb{R}^n$. Consequently, $(I + MM')|_{U^\perp} = I$ (by $A|_V$ we denote the restriction of an operator $A$ to a subspace $V$).

One can see that both $U$ and $U^\perp$ are invariant subspaces of $I + MM'$. If we choose bases in $U$ and in $U^\perp$ and then concatenate them, we get a basis of $\mathbb{R}^n$; in this basis the matrix of $I + MM'$ has the form

$$\begin{pmatrix} A & 0 \\ 0 & I \end{pmatrix} ,$$

where $A$ is the matrix of $(I + MM')|_U$. It remains to evaluate $\det(A)$.

First let us consider the case of linearly independent $x_1, x_2, \ldots, x_m$. They form a basis of $U$ and we may use it to calculate the determinant of the operator $(I + MM')|_U$. However,

$$(I + MM') x_i = x_i + \sum_{j=1}^{m} (x_j' x_i) x_j$$

and thus the matrix of the operator $(I + MM')|_U$ in the basis $x_1, x_2, \ldots, x_m$ is $I + M'M$.

The case of linearly dependent $x_1, x_2, \ldots, x_m$ follows by continuity. Indeed, $m$ vectors in an $n$-dimensional space with $n \geq m$ may be approximated by $m$ independent vectors to any degree of precision and the determinant is a continuous function of the elements of a matrix.

## Appendix B: Inverting a Partitioned Matrix

If a matrix $M$ is partitioned into

$$M = \begin{pmatrix} P & Q \\ R & S \end{pmatrix} ,$$

where $P$ and $Q$ are square matrixes, and $M^{-1}$ is partitioned in the same way into

$$M^{-1} = \begin{pmatrix} \widehat{P} & \widehat{Q} \\ \widehat{R} & \widehat{S} \end{pmatrix} ,$$

then

$$\widehat{P} = P^{-1} + P^{-1} Q \left( S - R P^{-1} Q \right)^{-1} R P^{-1},$$
$$\widehat{Q} = -P^{-1} Q \left( S - R P^{-1} Q \right)^{-1},$$
$$\widehat{R} = -\left( S - R P^{-1} Q \right)^{-1} R P^{-1},$$

and

$$\widehat{S} = \left( S - R P^{-1} Q \right)^{-1} R P^{-1} ,$$

provided all the inversions are possible (the formulae may be checked by direct calculation). To obtain (5), we apply these formulae to the $T \times T$ matrix $\widetilde{\mathbf{K}}$ from Protocol 3. We partition it in such a way as to obtain the $(T-1) \times (T-1)$ matrix $\mathbf{K}$ in the upper left corner.